\documentclass[letterpaper]{article}
\usepackage{aaai}
\usepackage{times}
\usepackage{helvet}
\usepackage{courier}
\usepackage{comment}
\usepackage{graphicx}
\usepackage{cite}
\usepackage{subfigure}
\usepackage{amsmath}
\usepackage{amssymb}
\usepackage{eurosym}
\usepackage{fancyhdr}
\usepackage{url}
\usepackage{xspace}
\usepackage{amsthm}
\usepackage[noend]{algpseudocode}
\usepackage{enumitem}
\usepackage{color}
\usepackage{url}
\usepackage{amsmath}
\usepackage{amssymb}
\usepackage{xspace}
\theoremstyle{definition}

\theoremstyle{remark}

\usepackage{color}
\usepackage{layouts}  
\usepackage{multirow}
\usepackage{rotating}  
\usepackage[top=2.0in, bottom=1.5in, left=1in, right=1in]{geometry}

\usepackage[bookmarks=false,colorlinks=true,linkcolor={blue},citecolor={blue},urlcolor={blue}]{hyperref}

\usepackage[italian,english]{babel} 
\usepackage{tabularx} 
\usepackage{amsmath} 
\usepackage{amssymb} 
\usepackage{stdclsdv} 
\usepackage[usenames,dvipsnames,svgnames,table]{xcolor} 
\usepackage{fancyhdr} 
\usepackage{caption} 

\usepackage{tikz}
\usepackage{xspace}
\usepackage{pgfplots}
\usepackage{sansmath}

\usepackage{pgfplotstable}
\usepackage{xtab}
\xentrystretch{-0.12}
\usepackage{array}
\usepackage{textcomp}

\usepackage{listings}
\usepackage{natbib}

\makeatletter
\def\copyright@year{}
\def\copyright@text{}
\def\copyright@on{}

\def\copyrighttext#1{}
\def\copyrightyear#1{}
\makeatother

\usepackage{amsmath}

\newcommand{\fct}{4,843,615\xspace}
\newcommand{\fctf}{4,731,718\xspace}
\newcommand{\nfct}{4,662,741\xspace}
\newcommand{\nf}{111,897\xspace}
\newcommand{\ndfsfeat}{236\xspace}

\newcommand{\dtrain}{2.663 million\xspace}
\newcommand{\dtune}{326K\xspace}
\newcommand{\dtest}{1.852 million\xspace}

\newcommand{\tffps}{214,705\xspace}
\newcommand{\tffns}{4607\xspace}
\newcommand{\tffpc}{190,821.51\xspace}
\newcommand{\tffnc}{686,626.40\xspace}
\newcommand{\tftotalc}{877,447\xspace}

\newcommand{\bbvafps}{289,124\xspace}
\newcommand{\bbvafns}{4741\xspace}
\newcommand{\bbvafpc}{319,421.93\xspace}
\newcommand{\bbvafnc}{125,138.24\xspace}
\newcommand{\bbvatotalc}{444,560\xspace}

\newcommand{\dfsprec}{0.22\xspace}
\newcommand{\dfsfscore}{0.35\xspace}
\newcommand{\dfsfps}{133,254\xspace}
\newcommand{\dfsfns}{4729\xspace}
\newcommand{\dfsfpc}{183,502.64\xspace}
\newcommand{\dfsfnc}{71,563.75\xspace}
\newcommand{\dfstotalc}{255,066\xspace}

\newcommand{\tffprNF}{8.9\%\xspace}
\newcommand{\tfprecNF}{0.187\xspace}
\newcommand{\tffscoreNF}{0.30\xspace}
\newcommand{\tffpsNF}{162,302\xspace}
\newcommand{\tffnsNF}{5061\xspace}
\newcommand{\tffpcNF}{96,139.09\xspace}
\newcommand{\tffncNF}{818,989.95\xspace}
\newcommand{\tftotalcNF}{915,129.05\xspace}

\newcommand{\bbvaprecNF}{0.1166\xspace}
\newcommand{\bbvafscoreNF}{0.20\xspace}
\newcommand{\bbvafpsNF}{289,124\xspace}
\newcommand{\bbvafnsNF}{4741\xspace}
\newcommand{\bbvafpcNF}{319,421.93\xspace}
\newcommand{\bbvafncNF}{125,138.24\xspace}

\newcommand{\dfsfprNF}{2.96\%\xspace}
\newcommand{\dfsprecNF}{0.41\xspace}
\newcommand{\dfsfscoreNF}{0.56\xspace}
\newcommand{\dfsfpsNF}{53,592\xspace}
\newcommand{\dfsfnsNF}{5247\xspace}
\newcommand{\dfsfpcNF}{39,341.88\xspace}
\newcommand{\dfsfncNF}{638,940.89\xspace}
\newcommand{\dfstotalcNF}{678,282.77\xspace}

\newcommand{\dfsprecNFSeven}{0.374\xspace}
\newcommand{\dfsfscoreNFSeven}{0.524\xspace}
\newcommand{\dfstotalcNFSeven}{735,229.05\xspace}

\newcommand{\dfsprecNFTwoOne}{0.359\xspace}
\newcommand{\dfsfscoreNFTwoOne}{0.511\xspace}
\newcommand{\dfstotalcNFTwoOne}{716,157.54\xspace}

\newcommand{\dfsprecNFThreeFive}{0.36\xspace}
\newcommand{\dfsfscoreNFThreeFive}{0.512\xspace}
\newcommand{\dfstotalcNFThreeFive}{675,854.12\xspace}

\newcommand{\dfsprecSeven}{0.223\xspace}
\newcommand{\dfsfscoreSeven}{0.356\xspace}
\newcommand{\dfstotalcSeven}{305,282.26\xspace}

\newcommand{\dfsprecTwoOne}{0.23\xspace}
\newcommand{\dfsfscoreTwoOne}{0.366\xspace}
\newcommand{\dfstotalcTwoOne}{314,590.34\xspace}

\newcommand{\dfsprecThreeFive}{0.236\xspace}
\newcommand{\dfsfscoreThreeFive}{0.373\xspace}
\newcommand{\dfstotalcThreeFive}{322,250.67\xspace}

\begin{document}

\title{Solving the ``false positives'' problem in fraud prediction \\{\LARGE Automated Data Science at an Industrial Scale}}
\date{}


\author{Roy Wedge \and James Max Kanter \and Kalyan Veeramachaneni\\
Data to AI Lab,\\
LIDS, MIT,\\
Cambridge, MA-02139
\AND
Santiago Moral Rubio \and Sergio Iglesias Perez\footnotemark[2]\\ Banco Bilbao Vizcaya Argentaria (BBVA)\\
Madrid, Spain}

\vspace{20mm}

\maketitle

\bgroup
\renewcommand{\thefootnote}{\fnsymbol{footnote}}
\footnotetext[2]{Authors are the founding members of the Computer Science and Artificial Intelligence Laboratory (CSAIL), MIT's Cybsersecurity initiative.}
\egroup

\section*{Abstract}
In this paper, we present an automated feature engineering based approach to dramatically reduce false positives in fraud prediction. False positives plague the fraud prediction industry. It is estimated that only 1 in 5 declared as fraud are actually fraud and roughly 1 in every 6 customers have had a valid transaction declined in the past year. To address this problem, we use the Deep Feature Synthesis algorithm to automatically derive behavioral features based on the  historical data of the card associated with a transaction. We generate 237 features ($>$100 behavioral patterns) for each transaction, and use a random forest to learn a classifier. We tested our machine learning model on data from a large multinational bank and compared it to their existing solution. On an unseen data of 1.852 million transactions, we were able to reduce the false positives by 54\% and provide a savings of 190K euros. We also assess how to deploy this solution, and whether it necessitates streaming computation for real time scoring. We found that our solution can maintain similar benefits even when historical features are computed once every 7 days.

\section{Introduction}

Digital payment systems are enjoying growing popularity, and the companies that run them now have the ability to store, as data, every interaction a customer has with a payment system. Combined, these two developments have led to a massive increase in the amount of available transaction data, and have made service providers better equipped than ever to handle the problem of fraud detection. 

Fraud detection problems are well-defined supervised learning problems, and data scientists have long been applying machine learning to help solve them \cite{brause1999neural,ghosh1994credit}. However, false positives still plague the industry \citep{javelin}. It is not uncommon to have false positive rates as high as 10-15\% and have only 1 in 5 transactions declared as fraud be truly fraud \citep{javelin}. These high rates present significant financial ramifications: many analysts have pointed out that false positives may be costing merchants more then fraud itself \footnote{https://blog.riskified.com/true-cost-declined-orders/}. 

To mitigate this, most enterprises have adopted a multi-step process that combines work by human analysts and machine learning models. This process usually starts with a machine learning model generating a risk score and combining it with expert-driven rules to sift out potentially fraudulent transactions. The resulting alerts pop up in a 24/7 monitoring center, where they are examined and diagnosed by human analysts, as shown in Figure~\ref{process}. This process can potentially reduce the false positive rate by 5\% -- but this improvement comes only with high (and very costly) levels of human involvement. Even with such systems in place, a large number of false positives remain.

In this paper, we present an improved machine learning solution to drastically reduce the ``false positives'' in the fraud prediction industry. Such a solution will not only have financial implications, but also reduce the alerts at the 24/7 control center, enabling security analysts to use their time more effectively, and thus delivering the true promise of machine learning/artificial intelligence technologies. 

We received a large, multi-year dataset from BBVA, containing 900 million transactions. We were also given fraud reports that identified a very small subset of transactions as fraudulent. Our task was to develop a machine learning solution that: (a) uses this rich transactional data in a transparent manner (no black box approaches), (b) competes with the solution currently in use by BBVA, and (c) is deployable, keeping in mind the real-time requirements placed on the prediction system. 

We would be remiss not to acknowledge the numerous machine learning solutions achieved by researchers and industry alike (more on this in Section~\ref{related}). However, the value of extracting patterns from historical data has been only recently recognized as an important factor in developing these solutions -- instead, the focus has generally been on finding the best possible model given a set of features, and even after this , few studies focused on extracting a handful of features.   Recognizing the importance of feature engineering \cite{domingos2012few} -- in this paper, we use an automated feature engineering approach to generate hundreds of features to exploit these large caches of data and  dramatically reduce the false positives.  

\section{Key findings and results}
 \noindent \textbf{Key to success is automated feature engineering}: Having access to rich information about cards and customers exponentially increases the number of possible features we can generate. However, coming up with ideas, manually writing software and extracting features can be time-consuming, and may require customization each time a new bank dataset is encountered. In this paper, we use an automated method called \textit{deep feature synthesis}(DFS) to rapidly generate a rich set of features that represent the patterns of use for a particular account/card. Examples of features generated by this approach are presented in Table~\ref{feature_examples}. 
 
 As per our assessment, because we were able to perform feature engineering  automatically \textit{via} Featuretools and machine learning tools, we were able to focus our efforts and time on understanding the domain, evaluating the machine learning solution for financial metrics ($>$60\% of our time), and communicating our results. We imagine tools like these will also enable others to focus on the real problems at hand, rather than becoming caught up in the mechanics of generating a machine learning solution.

\noindent \textbf{Deep feature synthesis obviates the need for streaming computing}: While the deep feature synthesis algorithm can generate rich and complex features using historical information, and these features achieve superior accuracy when put through machine learning, it still needs to be able to do this in real time in order to feed them to the model. In the commercial space, this has prompted the development of streaming computing solutions.

But, what if we could compute these features only once every $t$ days instead? During the training phase, the abstractions in deep feature synthesis allow features to be computed with such a ``\textit{delay}," and for their accuracy to be tested, all by setting a single parameter. For example, for a transaction that happened on August 24th, we could use features that had been generated on August 10th. If accuracy is maintained, the implication is that aggregate features need to be only computed once every few days, obviating the need for streaming computing. 
\\

\noindent \textbf{What did we achieve?}  

\textbf{DFS achieves a 91.4\% increase in precision compared to BBVA's current solution.} This comes out to a reduction of 155,870 false positives in our dataset -- a 54\% reduction. 

\textbf{The DFS-based solution saves 190K euros over \dtest transactions - a tiny fraction of the total transactional volume.} These savings are over \dtest transactions, only a tiny fraction of BBVA's yearly total, meaning that the true annual savings will be much larger.

\textbf{We can compute features, once every 35 days and still generate value} Even when DFS features are only calculated once every 35 days, we are still able to achieve an improvement of 91.4\% in precision. However, we do lose 67K euros due to approximation, thus only saving 123K total euros. This unique capability makes DFS is a practically viable solution. 

\textbf{}


\begin{table*}[htb!]

\centering
\begin{tabularx}{0.8\linewidth}{ r | >{\small\arraybackslash}l} 
\textbf{Information type} & \textbf{\normalsize Attribute recorded}\\ 
\hline
{\large Verification results} &   \\
{\small Card} & captures information about unique situations during card verification.  \\
{\small Terminal} & captures information about unique situations during verification at a terminal.  \\
\hline
{\large About the location} &   \\
{\small Terminal} & can print/display messages \\
 & can change data on the card  \\
 & maximum pin length it can accept  \\
 & serviced or not \\
 & how data is input into the terminal  \\
{\small Authentication} & device type  \\
 & mode  \\
\hline
{\large About the merchant} & unique id  \\
 & bank of the merchant  \\
 & type of merchant  \\
 & country  \\
\hline
{\large About the card} & authorizer  \\
\hline
{\large About the transaction} & amount  \\
 & timestamp  \\
 & currency  \\
 & presence of a customer  \\
\end{tabularx}
\caption{A transaction, represented by a number of attributes that detail every aspect of it. In this table, we are showing *only* a fraction of what is being recorded in addition to the $amount$, $timestamp$ and $currency$ for a transaction.  These range from whether the customer was present physically for the transaction to whether the terminal where the transaction happened was serviced recently or not. We categorize the available information into several categories.}\label{tab:transaction}
\end{table*}

\begin{table*}[htb]
\centering\small
\begin{tabular}{@{} ll @{}}
\hline
\textbf{Item} & \textbf{Number}\\ \hline
Cards & 7,114,018 \\ 
Transaction log entries& 903,696,131\\ 
Total fraud reports& 172,410\\ 
Fraudulent use of card number reports& 122,913\\
Fraudulent card reports matched to transaction& \nf \\
\hline
\end{tabular}
\vspace{2mm}
\caption{Overview of the data we use in this paper}
\label{dataoverview}
\end{table*}

\section{Related work}\label{related}

Fraud detection systems have existed since the late 1990s. Initially, a limited ability to capture, store and process data meant that these systems almost always relied on expert-driven rules. These rules generally checked for some basic attributes pertaining to the transaction -- for example, ``Is the transaction amount greater then a threshold?'' or ``Is the transaction happening in a different country?'' They were used to block transactions, and to seek confirmations from customers as to whether or not their accounts were being used correctly. 

Next, machine learning systems were developed to enhance the accuracy of these systems \cite{brause1999neural,ghosh1994credit}. Most of the work done in this area emphasized the modeling aspect of the data science endeavor -- that is, learning a \textit{classifier}. For example, \cite{chan1999distributed, stolfo1997credit} present multiple classifiers and their accuracy. Citing the non-disclosure agreement, they do not reveal the fields in the data or the features they created. Additionally, \cite{chan1999distributed} present a solution using only transactional features, as information about their data is unavailable. 

Starting with \cite{shen2007application}, researchers have started to create small sets of hand-crafted features, aggregating historical transactional information \cite{bhattacharyya2011data,panigrahi2009credit}. \cite{whitrow2009transaction} emphasize the importance of aggregate features in improving accuracy. In most of these studies, aggregate features are generated by aggregating transactional information from the immediate past of the transaction under consideration. These are features like ``number of transactions that happened on the same day'', or ``amount of time elapsed since the last transaction''. 

Fraud detection systems require instantaneous responses in order to be effective. This places limits on real-time computation, as well as on the amount of data that can be processed. To enable predictions within these limitations, the aggregate features used in these systems necessitate a streaming computational paradigm in production \footnote{https://mapr.com/blog/real-time-credit-card-fraud-detection-apache-spark-and-event-streaming/}, \footnote{https://www.research.ibm.com/foiling-financial-fraud.shtml} \cite{carcillo2017scarff}. As we will show in this paper, however, aggregate summaries of transactions that are as old as 35 days can provide similar precision to those generated from the most recent transactions, up to the night before. This poses an important question: When is streaming computing necessary for predictive systems? Could a comprehensive, automatic feature engineering method answer this question?


\section{Dataset}
Looking at a set of multiyear transactional data provided to us -- a snapshot of which is shown in Table~\ref{tab:transaction} -- a few characteristics stand out: 

\begin{figure*}[htb!]
    \centering
    \includegraphics[width=0.95\textwidth]{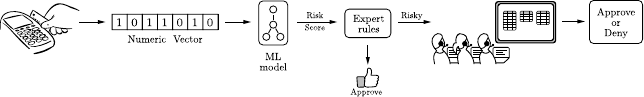}
    \caption{This graphic depicts the process of detecting and blocking fraudulent transactions in contemporary systems.}
    \label{process}
\end{figure*}

\begin{itemize}
    \item \textbf{Rich, extremely granular information}: Logs now contain not only information about a transaction's \textit{amount}, \textit{type}, \textit{time stamp} and \textit{location}, but also tangentially related material, such as the attributes of the terminal used to make the transaction. In addition, each of these attributes is divided into various subcategories that also give detailed information. Take, for example, the attribute that tells ``\textit{whether a terminal can print/display messages}''. Instead of a binary ``\textit{yes}'' or ``\textit{no},'' this attribute is further divided into multiple subcategories: ``\textit{can print}'', ``\textit{can print and display}'', \textit{``can display''}, ``\textit{cannot print or display}'', and \textit{``unknown''}. It takes a 59-page dictionary to describe each transaction attribute and all of its possible values. 
    \item \textbf{Historical information about card use}: Detailed, transaction-level information for each card and/or account is captured and stored at a central location, starting the moment the account is activated and stopping only when it is closed. This adds up quickly: for example, the dataset we received, which spanned roughly three years, contained 900 million transactions. Transactions from multiple cards or accounts belonging to the same user are now linked, providing a full profile of each customer's financial transactions.
\end{itemize}

\section{Data preparation}
Table~\ref{dataoverview} presents an overview of the data we used in this paper -- a total of 900 million transactions that took place over a period of 3 years. A typical transactional dataset is organized into a three-level hierarchy: $Customers \gets Cards \gets Transactions$. That is, a transaction belongs to a card, which belongs to a customer. Conversely, a card may have several transactions, and a customer may have multiple cards. This relational structure plays an important role in identifying subsamples and developing features. 

Before developing predictive models from the data, we took several preparative steps typical to any data-driven endeavor. Below, we present two data preparation challenges that we expect to be present across industry.

\noindent \textbf{Identifying a data subsample}:
 Out of the 900 million transactions in the dataset, only 122,000 were fraudulent. Thus, this data presents a challenge that is very common in fraud detection problems -- less then $0.002$\% of the transactions are fraudulent. To identify patterns pertaining to fraudulent transactions, we have to identify a subsample. Since we have only few examples of fraud, each transaction is an important training example, and so we choose to keep every transaction that is annotated as fraud.


However, our training set must also include a reasonable representation of the non-fraudulent transactions. We could begin by sampling randomly -- but the types of features we are attempting to extract also require historical information about the card and the customer to which a given transaction belongs. To enable the transfer of this information, we have to sample in the following manner:
\begin{enumerate}
\item Identify the cards associated with the fraudulent transactions,
\begin{itemize}
\item[--] Extract all transactions from these cards,
\end{itemize}
\item Randomly sample a set of cards that had no fraudulent transactions and, 
\begin{itemize}
\item[--] Extract all transactions from these cards. 
\end{itemize}
\end{enumerate}

Table~\ref{table:data} presents the sampled subset. We formed a training subset that has roughly 9.5 million transactions, out of which only \nf are fraudulent. These transactions give a complete view of roughly 72K cards. 

\noindent \textbf{Labeling transactions}: Fraud reports were collected from over one hundred daily logs. These logs did not include a $transactionID$ that would directly link them to transactions in the transaction log file, so we had to link them ourselves. Using the 
\begin{itemize}
\itemsep-0.1em
\item[--] \texttt{card number},
\item[--] \texttt{date of operation}, and
\item[--] \texttt{transaction amount} 
\end{itemize}
for each transaction listed in the fraud report, we attempted to match it to an entry in the transaction log file, comparing these numbers to
\begin{itemize}
\itemsep-0.1em
\item[--] \texttt{transaction amount}, 
\item[--] \texttt{date and time of operation},
\item[--] \texttt{original date and time}, and
\item[--] \texttt{currency used in every transaction made by that card}.
\end{itemize}
We formulated a set of rules that allowed us to pinpoint the IDs for fraudulent transactions. A total of 172,410 transactions were reported as fraud out of which 122,913 were fraudulent use of the card number. Out of these, with our matching rules, we were able to link a total of \nf transactions. 



\begin{table*}[htb!]
\centering
\begin{tabular}{ r >{\raggedright\arraybackslash}p{0.18\textwidth} >{\raggedright\arraybackslash}p{0.18\textwidth}>{\raggedright\arraybackslash}p{0.18\textwidth}}
\hline\\[-2ex]
& \textbf{Original Fraud} & \textbf{Non-Fraud} \\ \hline 
\# of Cards & 34378 & 36848 \\
\# of fraudulent transactions & \nf & 0 \\
\# of non-fraudulent transactions & \fctf& \nfct \\
\# of transactions & \fct & \nfct \\
\hline
\end{tabular}
\caption{The representative sample data set we extracted for training.}
\label{table:data}
\end{table*}%


\section{Automated feature generation}\label{afe}
Given the numerous attributes collected during every transaction, we can generate hypotheses/features in two ways:
\begin{itemize}
    \item[--] \textbf{By using only transaction information}: Each recorded transaction has a number of attributes that describe it, and we can extract multiple features from this information alone. Most features are binary, and can be thought of as answers to yes-or-no questions, along the lines of \textit{``Was the customer physically present at the time of transaction?''}. These features are generated by converting categorical variables using \texttt{one-hot-encoding}. Additionally, all the numeric attributes of the transaction are taken as-is. 
    \item[--] \textbf{By aggregating historical information}: Any given transaction is associated with a $card$, and we have access to all the historical transactions associated with that $card$. We can generate features by aggregating this information. These features are mostly numeric -- one example is, \textit{``What is the average amount of transactions for this card?''}. Extracting these features is complicated by the fact that, when generating features that describe a transaction at time $t$, one can only use aggregates generated about the $card$ using the transactions that took place \textit{before} $t$. This makes this process computationally expensive during the model training process, as well as when these features are put to use. 
\end{itemize} 

Broadly, this divides the features we can generate into two types: (a) so-called ``transactional features," which are generated from transactional attributes alone, and (b) features generated using historical data along with transactional features. Given the number of attributes and aggregation functions that could be applied, there are numerous potential options for both of these feature types. 

Our goal is to automatically generate numerous features and test whether they can predict fraud. To do this, we use an automatic feature synthesis algorithm called Deep Feature Synthesis \citep{kanter2015deep}. An implementation of the algorithm, along with numerous additional functionalities, is available as open source tool called \texttt{featuretools} \citep{featuretools}. We exploit many of the unique functionalities of this tool in order to to achieve three things: (a) a rich set of features, (b) a fraud model that achieves higher precision, and (c) approximate versions of the features that make it possible to deploy this solution, which we are able to create using a unique functionality provided by the library. In the next subsection, we describe the algorithm and its fundamental building blocks. We then present the types of features that it generated.

\subsection{Deep Feature Synthesis}
The purpose of Deep Feature Synthesis (DFS) is to automatically create new features for machine learning using the relational structure of the dataset. The relational structure of the data is exposed to DFS as \textit{entities} and \textit{relationships}.

An entity is a list of instances, and a collection of attributes that describe each one -- not unlike a table in a database. A transaction entity would consist of a set of transactions, along with the features that describe each transaction, such as the transaction amount, the time of transaction, etc.

A relationship describes how instances in two entities can be connected. For example, the point of sale (POS) data and the historical data can be thought of as a ``Transactions" entity and a "Cards" entity. Because each card can have many transactions, the relationship between Cards and Transactions can be described as a ``parent and child" relationship, in which each parent (Card) has one or more children (Transactions).

Given the relational structure, DFS searches a built-in set of \textit{primitive feature functions}, or simply called ``primitives", for the best ways to synthesize new features. Each primitive in the system is annotated with the data types it accepts as inputs and the data type it outputs. Using this information, DFS can stack multiple primitives to find \textit{deep features} that have the best predictive accuracy for a given problems. 

The primitive functions in DFS take two forms.

\begin{itemize}
    \item[--] Transform primitives: This type of primitive creates a new feature by applying a function to an existing column in a table. For example, the \texttt{Weekend} primitive could accept the transaction date column as input and output a columns indicating whether the transaction occurred on a weekend.
 \item[--] Aggregation primitives: This type of primitive uses the relations between rows in a table. In this dataset, the transactions are related by the id of the card that made them. To use this relationship, we might apply the \texttt{Sum} primitive to calculate the total amount spent to date by the card involved in the transaction.
\end{itemize}  
 
 \textbf{Synthesizing deep features}: For high value prediction problems, it is crucial to explore a large space of potentially meaningful features. DFS accomplishes this by applying a second primitive to the output of the first. For example, we might first apply the \texttt{Hour} transform primitive to determine when during the day a transaction was placed. Then we can apply \texttt{Mean} aggregation primitive to determine average hour of the day the card placed transactions. This would then read like \texttt{cards.Mean(Hour(transactions.date))} when it is auto-generated. If the card used in the transaction is typically only used at one time of the day, but the transaction under consideration was at a very different time, that might be a signal of fraud. 
 
 Following this process of stacking primitives, DFS enumerates many potential features that can be used for solving the problem of predicting credit card fraud. In the next section, we describe the features that DFS discovered and their impact on predictive accuracy. 
 
\begin{table*}
\begin{tabular}{r|l}
\hline 
\multicolumn{2}{c}{Features aggregating information from all the past transactions}\tabularnewline
Expression & Description\tabularnewline
\hline 
cards.MEAN(transactions.amount) & Mean of transaction amount\tabularnewline
cards.STD(transactions.amount) & Standard deviation of the transaction amount\tabularnewline
cards.AVG\_TIME\_BETWEEN(transactions.date) & Average time between subsequent transactions \tabularnewline
cards.NUM\_UNIQUE(transactions.DAY(date)) & Number of unique days \tabularnewline
cards.NUM\_UNIQUE(transactions.tradeid) & Number of unique merchants \tabularnewline
cards.NUM\_UNIQUE(transactions.mcc) & Number of unique merchant categories\tabularnewline
cards.NUM\_UNIQUE(transactions.acquirerid) & Number of unique acquirers \tabularnewline
cards.NUM\_UNIQUE(transactions.country) & Number of unique countries\tabularnewline
cards.NUM\_UNIQUE(transactions.currency) & Number of unique currencies \tabularnewline
\end{tabular}
\caption{Features generated using DFS primitives. Each feature aggregates data pertaining to past transactions from the card. The \textit{left} column shows how the feature is computed via. an expression. The \textit{right} column describes the feature in English. These features capture patterns in the transactions that belong to a particular card. For example, what was the mean value of the }\label{feature_examples}
\end{table*}

\begin{tabular}{r|cc}
\hline 
\multirow{2}{*}{Data used} & \multicolumn{2}{l}{Number of features}\tabularnewline
\cline{2-3} 
 & OHE  & Numeric\tabularnewline
\hline 
Transactional & 91 & 2\tabularnewline
Historical & 192 & 44\tabularnewline
\end{tabular}

\section{Modeling}
After the feature engineering step, we have \ndfsfeat features for \fct transactions. Out of these transactions, only \nf are labeled as fraudulent. With machine learning, our goal is to (a) learn a model that, given the features, can predict which transactions have this label, (b) evaluate the model and estimate its generalizable accuracy metric, and (c) identify the features most important for prediction. To achieve these three goals, we utilize a random forest classifier, which uses subsampling to learn multiple decision trees from the same data. 

\noindent \textbf{Learning the model}: We used \texttt{scikit-learn}'s classifier with 100 trees by setting \texttt{n\_estimators}=\texttt{100}, and used \texttt{class\_weight} = \texttt{'balanced'}. The ``balanced'' mode uses the values of $labels$ to automatically adjust weights so that they are inversely proportional to class frequencies in the input data.

\noindent \textbf{Identifying important features}: The random forests used in our model allow us to calculate the relative \textit{feature importances}. These are determined by calculating the average number of training examples the feature separated in the decision trees that used it. We use the model trained using all the training data and extract feature importances. 

\section{Evaluating the model}\label{emodel}
To enable comparison in terms of ``false positives'', we assess the model comprehensively. Our framework involves (a) meticulously splitting the data into multiple exclusive subsets, (b) evaluating the model for machine learning metrics, and (c) comparing it to two different baselines. Later, we evaluate the model in terms of the financial gains it will achieve (in Section~\ref{financial} ). 

\noindent \textbf{Machine learning metric}
To evaluate the model, we assessed several metrics, including the area under the receiver operating curve (AUC-ROC). Since non-fraudulent transactions outnumber fraudulent transactions 1000:1, we first pick the operating point on the ROC curve (and the corresponding threshold) such that the true positive rate for fraud detection is $>89\%$, and then assess the model's \texttt{precision}, which measures how many of the blocked transactions were in fact fraudulent. For the given true positive rate, the precision reveals what losses we will incur due to false positives.

\noindent \textbf{Data splits}: We first experiment with all the cards that had one or more fraudulent transactions. To evaluate the model, we split it into mutually exclusive subsets, while making sure that fraudulent transactions  are proportionally represented each time we split. We do the following:
\begin{itemize}
    \item[--] we first split the data into training and testing sets. We use 55\% of the data for training the model, called $D_{train}$, which amounts to approximately \dtrain transactions, 
    \item[--] we use an additional \dtune, called $D_{tune}$, to identify the threshold - which is part of the training process, 
    \item[--] the remaining \dtest million transactions are used for testing, noted as $D_{test}$. 
\end{itemize}

\noindent \textbf{Baselines}: 
We compare our model with two baselines.

\begin{itemize}
    \item[--]\textbf{Transactional features baseline}: In this baseline, we only use the fields that were available at the time of the transaction, and that are associated with it. We do not use any features that were generated using historical data via DFS. We use \texttt{one-hot-encoding} for categorical fields. A total of 93 features are generated in this way. We use a random forest classifier, with the same parameters as we laid out in the previous section.
    \item[--]\textbf{Current machine learning system at BBVA}: For this baseline, we acquired risk scores that were generated by the existing system that BBVA is currently using for fraud detection. We do not know the exact composition of the features involved, or the machine learning model. However, we know that the method uses only transactional data, and probably uses neural networks for classification.
\end{itemize}

\noindent \textbf{Evaluation process}:
\begin{itemize}
    \item[--] {Step 1:} Train the model using the training data - $D_{train}$.  
    \item[--] {Step 2:} Use the trained model to generate prediction probabilities, $P_{tu}$ for $D_{tune}$. 
    \item[--] {Step 3:} Use these prediction probabilities, and true labels $L_{tu}$ for $D_{tune}$ to identify the threshold. The threshold $\gamma$ is given by:
    \begin{equation}
       \gamma= \underset{\gamma}{\operatorname{argmax}}\; {\operatorname{precision}_\gamma} \times u\left({\operatorname{tpr}_\gamma}- 0.89\right)
    \end{equation}
    where $\texttt{tpr}_\gamma$ is the true positive rate that can be achieved at threshold $\gamma$ and $u$ is a unit step function whose value is 1 when $\texttt{tpr}_{\gamma} \geq 0.89$. The true positive rate ($tpr$) when threshold $\gamma$ is applied is given by: 
    \begin{equation}
        {\operatorname{tpr_\gamma}} = \frac{\sum\limits_i \delta(P_{tu}^i \geq \gamma)} {\sum\limits_i L_{tu}^i}, \forall i, \quad where \quad L_{tu}^i=1
    \end{equation}
    where $\delta(.)=1$ when $P_{tu}^i\geq \gamma$ and $0$ otherwise. Similarly, we can calculate $fpr_\gamma$ (false positive rate) and $precision_\gamma$.
    
    \item[--] {Step 4:} Use the trained model to generate predictions for $D_{test}$. Apply the threshold $\gamma$ and generate predictions. Evaluate \texttt{precision}, \texttt{recall} and \texttt{f-score}. Report these metrics. 
    
\end{itemize}


\begin{figure}[htb!]
    \centering
    \includegraphics[width=0.25\textwidth]{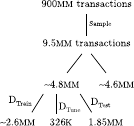}
    \caption{The process of splitting data into training /testing sets, followed by validation. Starting with approximately 9.5MM transactions, we split them into sets according to which cards had fraudulent transactions and which did not. For the \fct transactions from the cards that had fraudulent activity, we split them into three groups: \dtrain for training, \dtune for identifying threshold and \dtest for testing.}
    \label{datasplits}
\end{figure}





\begin{figure}[htb!]
    \centering
    \includegraphics[width=0.45\textwidth]{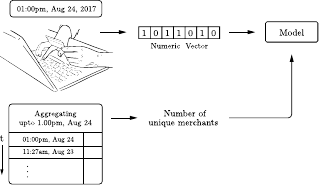}
    \caption{The process of using a trained model in real time. When a transaction is made, binary ``yes''/``no'' features are extracted from the transaction. Features that quantify patterns from the history of the card's transactions are extracted. For example, the feature ``\textit{Number of unique merchants this card had transactions with in the past}'' is extracted by identifying all the transactions for this card up to the current time and computing on it. Both of these feature sets are passed to the machine learning model to make predictions.}
    \label{dfs_feat}
\end{figure}

\begin{table}
\begin{tabular}{r|ccc}
Metric & Transactional & Current system & DFS\tabularnewline
\hline 
Precision &\tfprecNF  & \bbvaprecNF & \textbf{\dfsprecNF}\tabularnewline
F-Score &\tffscoreNF & \bbvafscoreNF & \textbf{\dfsfscoreNF }\tabularnewline
\end{tabular}
\caption{\texttt{Precision} and \texttt{f-score} achieved in detecting non-fraudulent transactions at the fixed \texttt{recall} (a.k.a true positive rate) of $>= 0.89$. We compare the performance of features generated using the deep feature synthesis algorithm to those generated by ``\texttt{one-hot-encoding''} of transactional attributes, and those generated by the baseline system currently being used. These baselines are described in Section~\ref{results}. A }\label{result_approximation}
\end{table}

\section{Results and discussion}\label{results}
\noindent \textbf{DFS solution}: In this solution we use the features generated by the DFS algorithm as implemented in \texttt{featuretools}. A total of 236 features are generated, which include those generated from the fields associated with the transaction itself. We then used a random forest classifier with the hyperparameter set described in the previous section.

In our case study, the transactional features baseline system has a false positive rate of \tffprNF, while the machine learning system with DFS features has a false positive rate of \dfsfprNF, a reduction of 6\%.

When we fixed the true positive rate at $>89\%$, our precision for the transactional features baseline was \tfprecNF. For the model that used DFS features, we got a precision of \dfsprecNF, a $>$2x increase over the baseline. When compared to the current system being used in practice, we got a $>$3x improvement in precision. The current system has a precision of only about \bbvaprecNF. 



\section{Financial evaluation of the model}\label{financial}
To assess the financial benefit of reducing false positives, we first detail the impact of false positives, and then evaluate the three solutions. When a false positive occurs, there is the possibility of losing a sale, as the customer is likely to abandon the item s/he was trying to buy. A compelling report published by Javelin Strategy \& Research reports that these blocked sales add up to \$118 billion, while the cost of real card fraud only amounts to \$9 billion \citep{javelin}. 
Additionally, the same \citep{javelin} study reports that 26\% of shoppers whose cards were declined reduced their shopping at that merchant following the decline, and 32\% stopped entirely. There are numerous other costs for the merchant when a customer is falsely declined \footnote{https://blog.riskified.com/true-cost-declined-orders/}.

From a card issuer perspective, when possibly authentic sales are blocked, two things can happen: the customer may try again, or may switch to a different issuer (different card). Thus issuers also lose out on millions in interchange fees, which are assessed at 1.75\% of every transaction.\footnote{"Interchange fee" is a term used in the payment card industry to describe a fee paid between banks for the acceptance of card-based transactions. For sales/services transactions, the merchant's bank (the "acquiring bank") pays the fee to a customer's bank (the "issuing bank").} Additionally, it also may cause customer retention problems. Hence, banks actively try to reduce the number of cards affected by false positives.

\begin{table*}
\centering
\begin{tabular}{r|llll>{\raggedright}m{2cm}}
\hline 
\multirow{2}{*}{Method } & \multicolumn{2}{l}{False postitives} & \multicolumn{2}{l}{False Negatives} & \multirow{2}{2cm}{Total Cost \euro}\tabularnewline
\cline{2-5} 
 & Number & Cost \euro & Number & Cost \euro & \tabularnewline
\hline 
Current system & \bbvafpsNF & \bbvafpcNF & \bbvafnsNF & \bbvafncNF & \bbvatotalc\tabularnewline

Transactional features only & \tffpsNF & \tffpcNF  & \tffnsNF  & \tffncNF  & \tftotalcNF \tabularnewline

DFS & \dfsfpsNF & \dfsfpcNF& \dfsfnsNF & \dfsfncNF & \dfstotalcNF \tabularnewline
\end{tabular}

\caption{Losses incurred due to false positives and false negatives. This table shows the results when \texttt{threshold} is tuned to achieve $tpr \geq 0.89$. \textbf{Method}: We aggregate the \texttt{amount} for each false positive and false negative. False negatives are the frauds that are not detected by the system. We assume the issuer fully reimburses this to the client. For false positives, we assume that 50\% of transactions will not happen using the card and apply a factor of 1.75\% for interchange fee to calculate losses. These are estimates for the validation dataset which contained approximately \dtest transactions.} \label{financial_gains_NF}
\end{table*}

To evaluate the financial implications of increasing the precision of fraud detection from \bbvaprecNF to \dfsprecNF, we do the following: 
\begin{itemize}
    \item[--] We first predict the \textit{label} for the \dtest transactions in our test dataset using the model and the threshold derived in Step 3 of the ``evaluation process''. Given the true \textit{label}, we then identify the transactions that are falsely labeled as frauds. 
    \item[--] We assess the financial value of the false positives by summing up the amount of each of the transactions (in Euros). 
    \item[--] Assuming that 50\% of these sales may successfully go through after the second try, we estimate the loss in sales using the issuer's card by multiplying the total sum by $0.5$. 
    \item[--] Finally, we assess the loss in interchange fees for the issuer at 1.75\% of the number in the previous step. This is the cost due to false positives - $cost_{fp}$
    \item[--] Throughout our analysis, we fixed the true positive rate at 89\%. To assess the losses incurred due to the remaining \~10\%, we sum up the total amount across all transactions that our model failed to detect as fraud. This is the cost due to false negatives - $cost_{fn}$
    \item[--] The total cost is given by 
    $$total cost = cost_{fp} + cost_{fn}$$
\end{itemize}

By doing the simple analysis as above, we found that our model generated using the DFS features was able to reduce the false positives significantly and was able to reduce the $cost_{fp}$ when compared to BBVA's solution (\euro \dfsfpcNF \textit{vs.} \euro \bbvafpcNF). But it did not perform better then BBVA overall in terms of the total $cost$, even though there was \textit{not} a significant difference in the number of false negatives between DFS based and BBVA's system. Table~\ref{financial_gains_NF} presents the detailed results when we used our current model as if. This meant that BBVA's current system does really well in detecting high valued fraud. To achieve similar effect in detection, we decided to re-tune the threshold. 

\noindent \textbf{Retuning the threshold}: To tune the threshold, we follow the similar procedure described in Section~\ref{emodel}, under subsection titled ``Evaluation process'', except for one change. In Step 2 we weight the probabilities generated by the model for a transaction by multiplying the amount of the transaction to it. Thus, 
\begin{equation}
    P_{tu}^i \gets P_{tu}^i \times amount^i 
\end{equation}

We then find the \texttt{threshold} in this new space. For test data, to make a decision we do a similar transformation of the probabilities predicted by a classifier for a transaction. We then apply the threshold to this new transformed values and make a decision. 
This weighting essentially reorders the the transactions. Two transactions both with the same prediction probability from the classifier, but vastly different amounts, can have different predictions. 

Table~\ref{financial_gains} presents the results when this new \texttt{threshold} is used. A few points are noteworthy:
\begin{itemize}
    \item[--] \textbf{DFS model reduces the total cost BBVA would incur by atleast 190K euros}. It should be noted that these set of transactions, \dtest, only represent a tiny fraction of overall volume of transactions in a year. We further intend to apply this model to larger dataset to fully evaluate its efficacy. 
    \item[--]\textbf{When threshold is tuned considering financial implications, precision drops}. Compared to the precision we were able to achieve previously, when we did not tune it for high valued transactions, we get less precision (that is more false positives). In order to save from high value fraud, our threshold gave up some false positives. 
    \item[--]\textbf{``Transactional features only'' solution has better precision than existing model, but smaller financial impact}: After tuning the threshold to weigh high valued transactions, the baseline that generates features only using attributes of the transaction (and no historical information) still has a higher precision than the existing model. However, it performs worse on high value transaction so the overall financial impact is the worse than BBVA's existing model.
    \item[--]\textbf{54\% reduction in the number of false positives}. Compared to the current BBVA solution, DFS based solution cuts the number of false positives by more than a half. Thus reduction in number of false positives reduces the number of cards that are false blocked - potentially improving customer satisfaction with BBVA cards.   
\end{itemize}

\begin{table*}
\centering
\begin{tabular}{r|llll>{\raggedright}m{2cm}}
\hline 
\multirow{2}{*}{Method } & \multicolumn{2}{l}{False postitives} & \multicolumn{2}{l}{False Negatives} & \multirow{2}{2cm}{Total Cost (Euros)}\tabularnewline
\cline{2-5} 
 & Number & Cost \euro & Number & Cost \euro & \tabularnewline
\hline 
Current system & \bbvafps & \bbvafpc & \bbvafns & \bbvafnc & \bbvatotalc\tabularnewline

Transactional features only & \tffps & \tffpc  & \tffns  & \tffnc  & \tftotalc \tabularnewline

DFS & \dfsfps & \dfsfpc& \dfsfns & \dfsfnc & \dfstotalc\tabularnewline
\end{tabular}

\caption{Losses incurred due to false positives and false negatives. This table shows the results when the \texttt{threshold} is tuned to consider high valued transactions. \textbf{Method}:We aggregate the \texttt{amount} for each false positive and false negative. False negatives are the frauds that are not detected by the system. We assume the issuer fully reimburses this to the client. For false positives, we assume that 50\% of transactions will not happen using the card and apply a factor of 1.75\% for interchange fee to calculate losses. These are estimates for the validation dataset which contained approximately \dtest transactions.} \label{financial_gains}
\end{table*}

\section{Real-time deployment considerations}\label{approx}
So far, we have shown how we can utilize complex features generated by DFS to improve predictive accuracy. Compared to the baseline and the current system, DFS-based features that utilize historical data improve the precision by 52\% while maintaining the recall at \~90\%.

However, if the predictive model is to be useful in real life, one important consideration is: how long does it take to compute these features in real time, so that they are calculated right when the transaction happens? This requires thinking about two important aspects:
\begin{itemize}
    \item[--] Throughput: This is the number of predictions sought per second, which varies according to the size of the client. It is not unusual for a large bank to request anywhere between 10-100 predictions per second from disparate locations.
    \item[--] Latency: This is the time between when a prediction is requested and when it is provided. Latency must be low, on the order of milliseconds. Delays cause annoyance for both the merchant and the end customer. 
\end{itemize}

\begin{figure*}[htb!]
    \centering
    \includegraphics[width=0.65\textwidth]{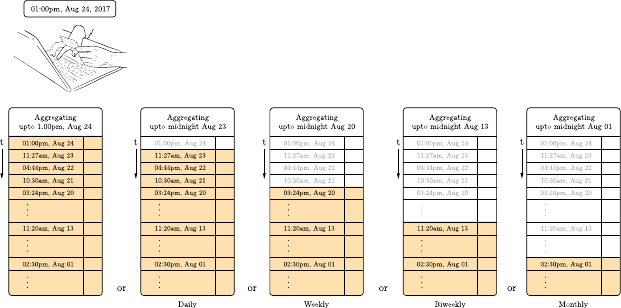}
    \caption{The process of approximating of feature values. For a transaction that happens at 1 PM on August 24, we can extract features by aggregating from transactions up to that time point, or by aggregating up to midnight of August 23, or midnight of August 20, and so on. Not shown here, these approximations implicitly impact how frequently the features need to be computed. In the first case, one has to compute the features in real time, but as we move from left to right, we go from computing on daily basis to once a month. }
    \label{approx}
\end{figure*}

While throughput is a function of how many requests can be executed in parallel as well as the time each request takes (the latency), latency is strictly a function of how much time it takes to do the necessary computation, make a prediction, and communicate the prediction to the end-point (either the terminal or an online or digital payment system). When compared to the previous system in practice, using the complex features computed with DFS adds the additional cost of computing features from historical data, on top of the existing costs of creating transactional features and executing the model. Features that capture the aggregate statistics can be computed in two different ways:
\begin{itemize}
    \item[--] \textbf{Use aggregates up to the point of transaction}: This requires having infrastructure in place to query and compute the features in near-real time, and would necessitate streaming computation. 
    \item[--] \textbf{Use aggregates computed a few time steps earlier}: We call these approximate features -- that is, they are the features that were current a few time steps $t$ ago. Thus, for a transaction happening at 1PM,  August 24, we could use features generated on August 1 (24 days old). This enables feature computation in a batch mode: we can compute features once every month, and store them in a database for every card. When making a prediction for a card, we query for the features for the corresponding card. Thus, the real-time latency is only affected by the query time. 
\end{itemize}

It is possible that using old aggregates could lead to a loss of accuracy. To see whether this would affect the quality of the predictions, we can simulate this type of feature extraction during the training process. \texttt{Featuretools} includes an option called \textit{approximate}, which allows us to specify the intervals at which features should be extracted before they are fed into the model. We can choose \texttt{approximate = "1 day"}, specifying that Featuretools should only aggregate features based on historical transactions on a daily basis, rather than all the way up to the time of transaction. We can change this to \texttt{approximate = "21 days"} or \texttt{approximate = "35 days"} \footnote{Detailed code examples using \texttt{approximate} are given in Appendix}. Figure~\ref{approx} illustrates the process of feature approximation.  To test how different metrics of accuracy are  effected -- in this case, the precision and the f1-score -- we tested for 4 different settings: \{1 day, 7 days, 21 days, and 35 days\}. 

Using this functionality greatly affects feature computation time during the model training process. By specifying a higher number of days, we can dramatically reduce the computation time needed for feature extraction. This enables data scientists to test their features quickly, to see whether they are predictive of the outcome.

Table~\ref{result_approximation_NF} presents the results of the approximation when \texttt{threshold} has been tuned to simply achieve $>0.89$ \texttt{tpr}. In this case, there is a loss of 0.05 in precision when we calculate features every 35 days.

In  Table~\ref{result_approximation} presents the \texttt{precision} and \texttt{f1-score} for different levels of approximation, when \texttt{threshold} is tuned taking the financial value of the transaction into account. Surprisingly, we note that even when we compute features once every 35 days, we do not loose any \texttt{precision}. However, we loose approximately $67K$ euros in money.


\textbf{Implications}: This result has powerful implications for our ability to deploy a highly precise predictive model generated using a rich set of features. It implies that the bank can compute the features for all cards once every 35 days, and still be able to achieve better accuracy then the baseline method that uses only transactional features. Arguably, a 0.05 increase in \texttt{precision}  as per Table~\ref{result_approximation_NF} and \euro 67K benefit as per Table~\ref{result_approximation} is worthwhile in some cases, but this should be considered alongside the costs it would incur to extract features on a daily basis. (It is also important to note that this alternative still only requires feature extraction on a daily basis, which is much less costly than real time.)

\section{Next steps}
Our next steps with the method developed in this paper are as follows:
\begin{itemize}
\item[--] Evaluate financial impact of the model on all 900 million transactions. 
\item[--] Measure impact of reducing false positives on customer retention. 
\item[--] Test the system performance on live data in production.
\end{itemize}

\section*{Acknowledgements}
BBVA authors would like to acknowledge their collaboration with the Computer Science and Artificial intelligence laboratory (CSAIL) alliances program. 

\section*{Open source acknowledgements}
This work would not have been possible without the open source software \texttt{Featuretools} and the time spent by software engineers at Feature Labs supporting the open source release. Likewise, acknowledgements are also due to open source software packages - \texttt{scikit-learn}, and \texttt{pandas}.

\begin{table*}
\begin{center}
\begin{tabular}{r|l l l l}
\multicolumn{1}{r|}{\multirow{2}{*}{Metric}} & \multicolumn{4}{c}{DFS with feature approximation} \\
\cline{2-5} 
 & 1 &  7 & 21  & 35\tabularnewline
\hline 
Precision  & \dfsprecNF & \dfsprecNFSeven & \dfsprecNFTwoOne   & \dfsprecNFThreeFive  \tabularnewline
F1-score & \dfsfscoreNF  &  \dfsfscoreNFSeven &   \dfsfscoreNFTwoOne   & \dfsfscoreNFThreeFive \tabularnewline

Total-cost &\dfstotalcNF &\dfstotalcNFSeven & \dfstotalcNFTwoOne & \dfstotalcNFThreeFive \tabularnewline
\end{tabular}
\end{center}
\caption{\texttt{Precision} and \texttt{f-score} achieved in detecting non-fraudulent transactions at the fixed \texttt{recall} (a.k.a true positive rate) of $>= 0.89$, when feature approximation is applied and \texttt{threshold} is tuned only to achieve a $tpr >=0.89$. A loss of 0.05 in precision is observed. No significant loss in financial value is noticed. }\label{result_approximation_NF}
\end{table*}

\begin{table*}
\begin{center}
\begin{tabular}{r|l l l l}
\multicolumn{1}{r|}{\multirow{2}{*}{Metric}} & \multicolumn{4}{c}{DFS with feature approximation} \\
\cline{2-5} 
 & 1 &  7 & 21  & 35\tabularnewline
\hline 
Precision  & \dfsprec & \dfsprecSeven & \dfsprecTwoOne   & \dfsprecThreeFive  \tabularnewline
F1-score & \dfsfscore  &  \dfsfscoreSeven &   \dfsfscoreTwoOne   & \dfsfscoreThreeFive \tabularnewline

Total-cost &\dfstotalc &\dfstotalcSeven & \dfstotalcTwoOne & \dfstotalcThreeFive \tabularnewline
\end{tabular}
\end{center}
\caption{\texttt{Precision} and \texttt{f-score} achieved in detecting non-fraudulent transactions at the fixed \texttt{recall} (a.k.a true positive rate) of $>= 0.89$, when feature approximation is applied and \texttt{threshold} is tuned to weigh high valued transactions more. No significant loss in \texttt{precision} is found, but an additional cost of approximately 67K euros is incurred. }\label{result_approximation}
\end{table*}

 \begin{table*}[htb!]
\centering
\begin{tabularx}{\linewidth}{@{}r|X@{}}
\hline
Feature primitive name & Description\\\hline
\multicolumn{2}{l}{Aggregation}\\ \hline
sum&sum of a numeric feature, or the number of ‘True’ values in a boolean feature\\ 
mean&a mean value of a numeric feature ignoring \\ 
std&finds the standard deviation of a numerical feature\\ 
count&counts the number of non-null values\\ 
number of unique&number of unique categorical variables\\ 
mode&most common element in a categorical feature\\ 
average time between&Maximum of absolute value of value minus previous value(diff)	\\ \hline
\multicolumn{2}{l}{Transformation}\\ \hline
weekend&transform datetime feature into boolean of weekend\\ 
day&transform Datetime feature into the day (0 - 30) of the month, or Timedelta features into number of days they encompass\\ \hline
\end{tabularx}
\vspace{2mm}
\caption{Numerical and Categorical \textit{feature functions}}
\label{table:feature_functions}
\end{table*}

\bibliographystyle{aaai}
\bibliography{bibliography}

\end{document}